\documentclass[letterpaper]{article} 
\usepackage{aaai2026}
\usepackage{times}  
\usepackage{helvet}  
\usepackage{courier}  
\usepackage[hyphens]{url}  
\usepackage{graphicx} 
\usepackage{natbib}  
\usepackage{caption} 
\usepackage{algorithm}
\usepackage{algorithmic}

\usepackage{newfloat}
\usepackage{listings}

\usepackage{makecell}
\usepackage{graphicx}
\usepackage{multirow}
\usepackage[table]{xcolor}
\usepackage[dvipsnames]{xcolor}
\usepackage{booktabs}
\usepackage{pifont}
\usepackage{amsmath}
\usepackage{stix}
\usepackage{xcolor}
\usepackage{colortbl}

\DeclareCaptionStyle{ruled}{labelfont=normalfont,labelsep=colon,strut=off} 
\floatstyle{ruled}
\newfloat{listing}{tb}{lst}{}
\floatname{listing}{Listing}
\newcommand{\gc}[1]{\cellcolor{lightgray!50}#1}

\def\model{qa-FLoRA}

\title{\model{}: Data-free query-adaptive Fusion of LoRAs for LLMs}

\author{
Shreya Shukla,~Aditya Sriram,~Milinda Kuppur Narayanaswamy,~Hiteshi Jain
}
\affiliations{
    Mercedes Benz Research and Development India \\
    \{shreya.shukla, aditya.a.sriram, milinda.narayana\_swamy, hiteshi.jain\}@mercedes-benz.com
}

\begin{document}

\maketitle

\begin{abstract}
The deployment of large language models for specialized tasks often requires domain-specific parameter-efficient finetuning through Low-Rank Adaptation (LoRA) modules. However, effectively fusing these adapters to handle complex, multi-domain composite queries remains a critical challenge. 
Existing LoRA fusion approaches either use static weights, which assign equal relevance to each participating LoRA, or require data-intensive supervised training for every possible LoRA combination to obtain respective optimal fusion weights. We propose \model{}, a novel query-adaptive data-and-training-free method for LoRA fusion that dynamically computes layer-level fusion weights by measuring distributional divergence between the base model and respective adapters. 
Our approach eliminates the need for composite training data or domain-representative samples, making it readily applicable to existing adapter collections. Extensive experiments across nine multilingual composite tasks spanning mathematics, coding, and medical domains, show that \model{} outperforms static fusion by $\sim$5\% with LLaMA-2 and $\sim$6\% with LLaMA-3, and the training-free baselines by $\sim$7\% with LLaMA-2 and $\sim$10\% with LLaMA-3, while significantly closing the gap with supervised baselines. Further, layer-level analysis of our fusion weights reveals interpretable fusion patterns, 
demonstrating the effectiveness of our approach for robust multi-domain adaptation.
\end{abstract}


\section{Introduction}

Large language models (LLMs) have demonstrated remarkable capabilities across a wide range of tasks, but their deployment to unseen or specialized tasks often requires domain-specific fine-tuning. However, standard full fine-tuning of an LLM is resource-intensive and can lead to catastrophic forgetting~\cite{luo2023forgetting}. Low-Rank Adaptation (LoRA)~\cite{hu2022lora} has emerged as a parameter-efficient fine-tuning technique that uses a low-rank approximation of the parameter update matrices to reduce the effective number of trainable parameters. LoRA’s low-rank updates effectively act as plug-and-play modules, i.e., once a LoRA adapter is trained for a particular task, it can be loaded into the base LLM at inference time without modifying the original parameters. Consequently, the same pre-trained base model can be reused across multiple downstream tasks by simply swapping in the appropriate LoRA modules. However, relying on individual LoRA modules in isolation fundamentally limits the model’s ability to handle complex or composite inputs that span multiple domains or tasks. In such scenarios, training dedicated adapters for every possible task combination is impractical and does not scale well with the combinatorial explosion of domains and tasks.

\begin{table}[t!]
\small
\centering
\resizebox{1.0\columnwidth}{!}{%
\begin{tabular}{l c c c c}
\toprule
\textbf{Method} & \makecell{\bf Query \\ \bf Adaptive} & \makecell{\bf Data \\ \bf required} & \makecell{\bf Supervised \\ \bf training} & \makecell{ \bf Per \\ \bf layer 
} \\
\midrule
Static Fusion & \color{red}{\ding{55}} & \color{Green}{\ding{55}} & \color{Green}{\ding{55}} & \color{red}{\ding{55}} \\

LoraFlow~\cite{wang2024loraflow} & \color{Green}{\ding{51}} & \color{red}{\ding{51}} & \color{red}{\ding{51}} & \color{Green}{\ding{51}} \\

LoraHub~\cite{huang2023lorahub} & \color{Green}{\ding{51}} & \color{red}{\ding{51}} & \color{red}{\ding{51}} & \color{red}{\ding{55}} \\

Centroid Sim.~\cite{belofsky2023_cosine} & \color{Green}{\ding{51}} & \color{red}{\ding{51}} & \color{Green}{\ding{55}} & \color{red}{\ding{55}} \\
\midrule
\textbf{\model{}(Ours)} & \color{Green}{\ding{51}} & \color{Green}{\ding{55}} & \color{Green}{\ding{55}} & \color{Green}{\ding{51}} \\
\bottomrule
\end{tabular}
}
\caption{Comparison of existing LoRA fusion approaches. ({\color{red}$\smblksquare$}) indicates an undesired trait, ({\color{Green}$\smblksquare$}) indicates a desired one.}
\label{tab:intro_figure}
\end{table}

This challenge has motivated the growing body of research on LoRA fusion, which aims to integrate multiple task-specific adapters to enable robust inference across composite inputs spanning diverse domains. 
Early works relied on \textit{static merging}~\cite{model-merging}, which naively combines adapters with fixed weights. This method does not account for the semantic relevance of domain experts to individual queries. More recent \textit{supervised approaches} adopt dynamic fusion schemes inspired by the Mixture-of-Experts(MoE) architecture~\cite{jiang2024mixtralexperts}, and train a routing network to predict fusion weights~\cite{wang2024loraflow, xu2024meteora}. While these dynamic fusion methods improve adaptability to individual queries, they still require re-training the router for each new adapter or domain addition. Moreover, such training-based methods require a diverse collection of composite training data for all possible adapter combinations, 
thus creating a scalability bottleneck that limits their applicability to heterogeneous adapter collections.

To address the scalability limitations of training-based approaches, another line of work has explored \textit{training-free} LoRA fusion that bypasses the requirement of composite data for fusion weights optimization. These methods typically compute fusion weights by measuring cosine similarity between test queries and precomputed domain-centroids 
for each adapter~\cite{belofsky2023_cosine, chronopoulou2023adaptersoup}. However, the effectiveness of such centroid-based approaches is highly dependent on the quality and representativeness of the domain-specific data used to compute centroids. Moreover, this method fails to capture the distributional shifts that adapters induce at different layers of the LLM, and for domains with semantically similar representations, centroids provide inadequate information, leading to suboptimal fusion weights. Table~\ref{tab:intro_figure} compares the pros and cons of the existing methods. These challenges highlight the need for a more flexible and robust training-free method for dynamic fusion of LoRA adapters. 

In this paper, we introduce \model{}, a novel data-and-training-free method that can dynamically determine layer-level weights for query-adaptive fusion of LoRA modules. 
Our approach is grounded in the following insight -- examining how each adapter modifies the base model's predictions reveals its relevance to the query. Specifically, when a LoRA adapter is semantically relevant to an input, it injects meaningful task-specific information that diverges from the base model's representation in a measurable way. This divergence serves as a proxy for semantic relevance, enabling dynamic weighing of adapters based on their contribution to the query at hand. Notably, our proposed approach eliminates the need for composite training data or domain-specific representative samples as required by previous approaches. \\

\noindent The key contributions of our work are threefold: 
\begin{enumerate}
    \item We propose \model{}, a novel data-free and training-free approach for query-adaptive LoRA Fusion that dynamically computes layer-level fusion weights based on the semantic relevance of adapters to individual queries.
    \item We extensively compare our method with diverse baselines across static, supervised, and training-free fusion paradigms. We demonstrate substantial improvements over static and training-free methods (by 5\% and 7\% with LLaMA-2-7B and by 6\% and 10\% with LLaMA-3-8B base LLM respectively), while significantly closing the performance gap with fully supervised methods.
    \item Through comprehensive evaluation across nine different composite tasks, we validate that our approach can effectively combine diverse domain expertise without requiring additional training, making it readily applicable to existing LoRA adapter collections.
\end{enumerate}

\section{Related Work}

\noindent \textbf{Parameter-Efficient Fine-Tuning (PEFT) of LLMs.} 
With the recent advances in PEFT techniques~\cite{han2024peft_survey_1, xu2023peft_survey_2}, LLMs are often domain-adapted by either updating only a small subset of model parameters or adding lightweight task-specific trainable modules. Existing PEFT strategies can broadly be classified into three: \textit{additive methods}~\cite{houlsby2019additive_1, he2021additive_2, zhu2021additive_3, lei2023additive_4, chen2023_additive_5} that introduce new trainable modules, \textit{reparameterization methods}~\cite{hu2022lora, valipour2022dylora, zhang2023adalora, zhang2023increlora, hayou2024lora+, liu2024dora} that express updates using low-rank adaptation of the parameter update matrix, and \textit{selective methods}~\cite{guo2020_selective_parameter, zaken2021_selective_bitfit, sung2021_selective_training, he2022_selective_sparseadapter, das2023_selective_unified, liao2023_selective_parameter, zhang2023_selective_loraprune} that fine-tune only chosen existing weights.
In this work, our focus is on reparameterization methods, particularly LoRA~\cite{hu2022lora}.

\noindent \textbf{LoRA Fusion for multi-task adaptation.} 
LoRA fusion combines multiple domain-experts (LoRA modules) to enable robust inference across multi-domain composite inputs.
The simplest approach to combining multiple adapters is static LoRA fusion, which uses arithmetic operations (averaging, weighted averaging, or task arithmetic) to merge adapters offline~\cite{model-merging}. This method 
fails to adapt to the varying semantic requirements of input queries, resulting in suboptimal performance.
Existing methods for dynamic LoRA fusion predominantly rely on supervised learning to train routing mechanisms~\cite{zadouri2023moe, kong2024loraswitch, luo2024moelora, ma2024modula}. LoraRetriever~\cite{zhao2024loraretriever} combines retrieval-based selection with composition strategies. LoRAMoE~\cite{dou2023loramoe} utilizes mixture-of-experts gating networks for token-level adapter selection. DLP-LoRA~\cite{zhang2024dlplora} proposes lightweight plugins and dynamic merging strategies for multi-task scenarios. LoRA-Flow~\cite{wang2024loraflow} introduces progressive fusion with learnable gates, and MeteoRA~\cite{xu2024meteora} implements token-level gating for fine-grained control. Another work LoRAHub~\cite{huang2023lorahub} employs gradient-free few-shot optimization to learn fusion weights in a non-parametric fashion.
However, the above supervised methods require composite training data for all possible adapter combinations, to optimize fusion weights, which limits their generalizability to unseen task combinations. 
Existing training-free approaches~\cite{belofsky2023_cosine, chronopoulou2023adaptersoup} rely on cosine similarity between test queries and pre-computed centroids of domain-specific data to select relevant adapters. 
However, there is still a dependency on domain-specific data for centroid computation, and the per-layer distributional shifts are not taken into account.
To address these limitations, we propose \model{}, a query-adaptive data-and-training-free LoRA-Fusion method that leverages divergence between the base model and adapter distributions to dynamically identify the most semantically relevant adapters, without requiring additional parametric routing or few-shot data.

\section{Our Approach}

\begin{figure*}[!t]
\centering
\includegraphics[width=1.0\textwidth]{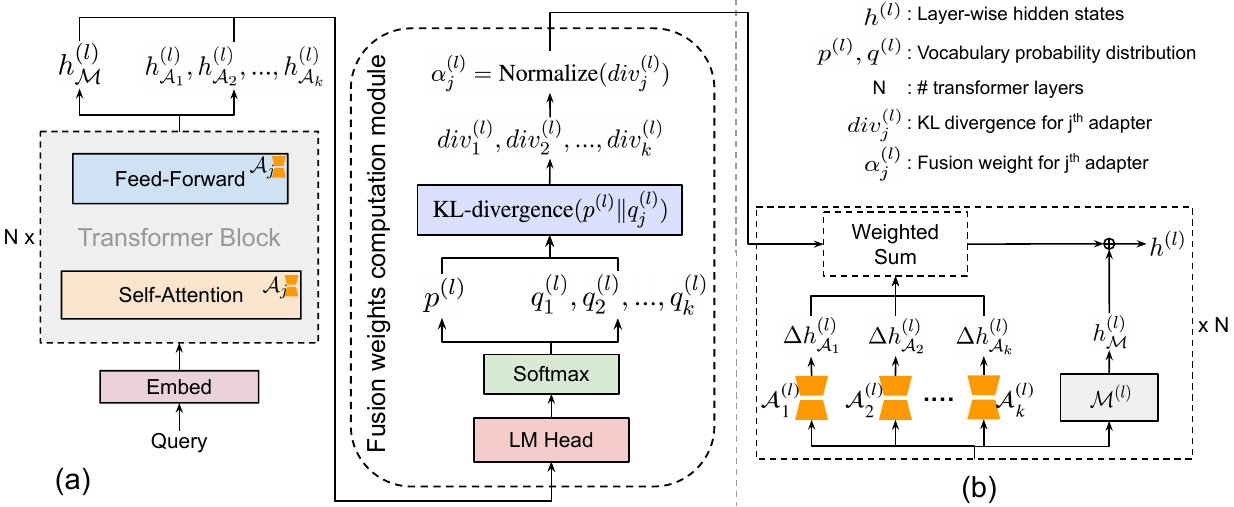}
\caption{\textbf{Proposed \model{} framework.} For an input query, we \textbf{(a)} dynamically calculate the per-layer fusion weights by utilizing the KL divergence between base model and adapter vocabulary distributions, and \textbf{(b)} perform weighted combination of LoRA adapter outputs with the base model for every transformer layer of the LLM.}
\label{fig:QA-LoRAF}
\end{figure*}

In this section, we present \model{}, a novel data-and-training-free approach for \underline{\textbf{q}}uery-\underline{\textbf{a}}daptive \underline{\textbf{F}}usion of \underline{\textbf{LoRA}} modules, that leverages the distributional divergence of each adapter with respect to the base LLM, to identify the semantic relevance of adapters for each input query. Figure~\ref{fig:QA-LoRAF} illustrates the overall framework of our proposed approach. \\

\noindent \textbf{Problem Formulation} \\
Given a frozen large language model $\mathcal{M}$ with parameters $W$ and a set of $k$ domain-specific LoRA adapters $\{\mathcal{A}_1, \mathcal{A}_2, \ldots, \mathcal{A}_j, \ldots, \mathcal{A}_k\}$, each of which induces a low‐rank update $\Delta W_j$ to $W$. For an input query $Q$, our objective is to dynamically determine the per-layer fusion weights $\{\alpha^{(1)}_j, \alpha^{(2)}_j, \ldots, \alpha^{(l)}_j, \ldots, \alpha^{(N)}_j\}$ for an adapter $\mathcal{A}_j$ when computing the model predictions.

To achieve this, we (1) compute the layer-wise probability distributions for both the base model and each LoRA adapter as described in section~\ref{subsec: vocab_dist}, (2) quantify the distributional divergence between adapters and the base model to derive adapter fusion weights, as described in section~\ref{subsec: fusion_weights} and (3) perform weighted LoRA fusion with the base model to compute final predictions, as described in section~\ref{subsec: lora_fusion}.

\subsection{Layer-level probability distribution}
\label{subsec: vocab_dist}
This stage involves extracting intermediate hidden-state representations from both the base model and the adapters, and projecting their logits to vocabulary space to enable meaningful distributional comparisons of the base model and the adapters. \\

\subsubsection{Extraction of layer-level hidden states.}
For an input query $Q$, we process it through the base LLM $\mathcal{M}$ to obtain the layer-wise hidden states as $\textbf{h}^{(l)}_\mathcal{M} = W^{(l)}\textbf{h}^{(l-1)}_{\mathcal{M}}$, where $W^{(l)}$ denotes the weights of $l^{th}$ transformer layer of the base LLM $\mathcal{M}$. Similarly, we obtain the hidden states when processing the query $Q$ through each of the $k$ LoRA adapters as 
$\textbf{h}^{(l)}_{\mathcal{A}_{j}} = \textbf{h}^{(l)}_\mathcal{M} + \Delta W^{(l)}_{j}\textbf{h}^{(l-1)}_{\mathcal{A}_{j}}$. Here, for $l$=1, $\textbf{h}^{(l-1)}_{\mathcal{M}} = \textbf{h}^{(l-1)}_{\mathcal{A}_{j}} = x$, where $x$ denotes the query embeddings. For brevity and consistency, we talk about $\textbf{h}^{(l)}$ and $W^{(l)}$ at the transformer block level. The actual computations happen at linear-layer level for self-attention and feedforward networks within each transformer block. \\

\subsubsection{Projection onto vocabulary distribution.}  
To compute meaningful divergences between layer‐level representations, we must first project each hidden state $\textbf{h}^{(l)}$ onto the model’s vocabulary space. Notably, we reuse the pre‐trained LM head parameterized by $W_{LM}$ to produce logits for every layer as 
$\textbf{z}^{(l)}_\mathcal{M} = W_{LM}\textbf{h}^{(l)}_\mathcal{M}$ and $\textbf{z}^{(l)}_{\mathcal{A}_j} = W_{LM}\textbf{h}^{(l)}_{\mathcal{A}_j}$. The LM head is originally trained to process only the final‐layer hidden states. However, similar to~\cite{kavehzadeh2023lm_head_1, varshney2023lm_head_2}, we empirically found that applying the same projection to intermediate hidden-states yields well‐calibrated logits for divergence computation. 

Finally, we convert these logits into probability distribution over the vocabulary by applying softmax normalization
$$p^{(l)} = \frac{\exp(\textbf{z}^{(l)}_{\mathcal{M}})}{\sum_{m=1}^{d} \exp(\textbf{z}^{(l)}_{\mathcal{M}(m)})} 
\text{  ;  } 
q^{(l)}_{j} = \frac{\exp(\textbf{z}^{(l)}_{\mathcal{A}_j})}{\sum_{m=1}^{d} \exp(\textbf{z}^{(l)}_{\mathcal{A}_j(m)})}$$
to obtain $p^{(l)}$ and $q^{(l)}_{j}$ which denote the probability distribution of the outputs of layer $l$ from base LLM and $j_{th}$ adapter respectively. $d$ denotes the dimensionality of the logits.

\subsection{Distributional divergence and fusion weights}
\label{subsec: fusion_weights}
Here, we quantify how the predictions of each LoRA adapter diverge from the predictions of the base model. As shown in Figure~\ref{fig:QA-LoRAF}(a), for each layer $l$, we obtain the respective hidden state probability distributions for the last token of the query $Q$, and compute the Kullback Leibler (KL) divergence between the distribution of the base LLM $p^{(l)}$[-1] and each adapter $q^{(l)}_{j}$[-1] as shown in equation~\ref{eq:kl_div}.
\begin{equation}
    div^{(l)}_{j}(Q, \mathcal{A}_j) = D_{KL}(p^{(l)}\text{[-1]} \| q^{(l)}_{j}\text{[-1]})
\label{eq:kl_div}
\end{equation}
where:
\begin{equation}
    D_{KL}(p^{(l)} \| q^{(l)}_{j}) = \sum_{i=1}^{d} p^{(l)}_{(i)} \log \frac{p^{(l)}_{(i)}}{q^{(l)}_{j(i)}}
\label{eq:kl_div_formula}
\end{equation}
$d$ is the dimensionality of the probability distributions.

Intuitively, for a given query, the KL divergence $D_{KL}(p^{(l)} \| q^{(l)}_{j})$ measures the information gain when using the adapter distribution $q_{j}$ instead of the base model distribution $p$, thus quantifying the semantic information injected by each LoRA adapter relative to the base model representation. A higher KL divergence value indicates that the respective adapter is contributing task-specific information that the base model alone does not capture. Conversely, a lower KL divergence implies that the adapter provides little additional semantic value for the given query.

Once the KL divergence between the respective probability distributions is computed, the LoRA fusion weights for adapter $\mathcal{A}_{j}$ at each transformer layer can be obtained as 
$$\alpha^{(l)}_{j} = \frac{div^{(l)}_{j}}{\sum_{i=1}^{k} div^{(l)}_i}$$

\subsection{Adaptive LoRA fusion}
\label{subsec: lora_fusion}
As shown in Figure~\ref{fig:QA-LoRAF}(b), we fuse the LoRA adapters with respective per-layer fusion weights $\{\alpha^{(1)}_1, \ldots, \alpha^{(N)}_1\}, \ldots, \{\alpha^{(1)}_k, \ldots, \alpha^{(N)}_k\}$, and obtain the final model predictions as shown in equation~\ref{eq:lora_combi}.
\begin{equation}
    O = O_{\mathcal{M}} + \Delta O_{\mathcal{A}_{j}} =  (W + \sum_{j=1}^{k} \alpha_{j} \Delta W_{j})x
\label{eq:lora_combi}
\end{equation}
This per-layer adaptive fusion mechanism ensures that for each input query, the most semantically relevant adapters receive higher weights while the less relevant ones are naturally downweighted, enabling the model to dynamically and effectively combine diverse domain expertise for improved performance across heterogeneous tasks, without requiring additional training or optimization.

\label{tab:main_results}
\begin{table*}[t!]
\centering
\resizebox{1.0\textwidth}{!}{
\scriptsize
\begin{tabular}{l|l|l|ccc|c|ccc|c|ccc|c|c}
\toprule

 \multirow{2}{*}{\textbf{Base LLM}} & \multirow{2}{*}{\textbf{Paradigm}} & \multirow{2}{*}{\textbf{Method}} & \multicolumn{4}{c|}{\textbf{Math (accuracy)}} & \multicolumn{4}{c|}{\textbf{Code (pass@1)}} & \multicolumn{4}{c|}{\textbf{Medical (accuracy)}} & \multirow{2}{*}{\makecell{\textbf{Avg across} \\ \textbf{3 domains}}} \\
 
\cmidrule(lr){4-7} \cmidrule(lr){8-11} \cmidrule(lr){12-15}

 & & & \textbf{zh} & \textbf{ru} & \textbf{es} & \textbf{Avg} & \textbf{zh} & \textbf{ru} & \textbf{es} & \textbf{Avg} & \textbf{zh} & \textbf{ru} & \textbf{es} & \textbf{Avg} & \\
\midrule

\multirow{6}{*}{LLaMA-2-7B}
 & Static fusion & Avg [0.5, 0.5] & 12.8 & 10.4 & 18.4 & 13.9 & 17.1 & 17.7 & 18.3 & 17.7 & 28.0 & 33.0 & 28.0 & 29.7 & 20.4 \\
\cmidrule(lr){2-16}

 & \multirow{2}{*}{Supervised}
 & LoRAFlow & 33.2 & 37.6 & 42.0 & 37.6 & 20.7 & 23.8 & 23.2 & 22.6 & 31.7 & 35.3 & 30.6 & 32.5 & 30.9 \\
 & & LoRAHub & 20.8 & 28.4 & 36.8 & 28.7 & 19.5 & 21.3 & 20.1 & 20.3 & 30.5 & 33.2 & 26.7 & 30.1 & 26.4 \\
\cmidrule(lr){2-16}

 & \gc{Training free} & \gc{Centroid sim.} & \gc{8.4} & \gc{4.4} & \gc{17.6} & \gc{10.1} & \gc{\textbf{21.7}} & \gc{\textbf{16.5}} & \gc{\textbf{18.3}} & \gc{\textbf{18.8}} & \gc{\textbf{32.4}} & \gc{32.7} & \gc{17} & \gc{27.4} & \gc{18.8} \\
 & \gc{Data \& Training free} & \gc{\textbf{\model{} (Ours)}} & \gc{\textbf{21.6}} & \gc{\textbf{21.6}} & \gc{\textbf{36.4}} & \gc{\textbf{26.5}} & \gc{20.9} & \gc{\textbf{16.5}} & \gc{15.6} & \gc{17.7} & \gc{30.0} & \gc{\textbf{39.0}} & \gc{\textbf{31.0}} & \gc{\textbf{33.3}} & \gc{\textbf{25.8}} \\
\midrule

\multirow{5}{*}{LLaMA-3-8B}
 & Static fusion & Avg [0.5, 0.5] & 40.8 & 45.2 & 49.2 & 45.1 & 48.2 & 23.8 & 22.6 & 31.5 & 42.4 & 40.0 & 34.7 & 39.0 & 38.5 \\
\cmidrule(lr){2-16}

 & Supervised & LoRAFlow & 56.8 & 60.4 & 69.2 & 62.1 & 36.6 & 28.7 & 37.2 & 34.2 & 43.2 & 39.3 & 43.5 & 42.0 & 46.1 \\
\cmidrule(lr){2-16}

 & \gc{Training free} & \gc{Centroid sim.} & \gc{34.4} & \gc{41.6} & \gc{45.6} & \gc{40.5} & \gc{43.9} & \gc{\textbf{28.7}} & \gc{27.4} & \gc{33.3} & \gc{35.3} & \gc{22.7} & \gc{30.6} & \gc{29.5} & \gc{34.4} \\
 & \gc{Data \& Training free}& \gc{\textbf{\model{} (Ours)}} & \gc{\textbf{50.4}} & \gc{\textbf{58.4}} & \gc{\textbf{66.0}} & \gc{\textbf{58.3}} & \gc{\textbf{48.2}} & \gc{23.8} & \gc{\textbf{31.1}} & \gc{\textbf{34.4}} & \gc{\textbf{39.6}} & \gc{\textbf{38.0}} & \gc{\textbf{42.2}} & \gc{\textbf{39.9}} & \gc{\textbf{44.2}} \\

\bottomrule
\end{tabular}}
\caption{Quantitative comparison of different fusion methods across nine composite tasks with LLaMA-2-7B and LLaMA-3-8B as base LLMs. Best results within the training-free paradigm are highlighted in bold.}
\label{tab:main_results}
\end{table*}

\section{Experiments and Results}

\subsection{Setup}
\label{subsec:setup}
Our experiments are constrained to LLaMA-2-7B~\cite{touvron2023llama} and LLaMA-3-8B~\cite{grattafiori2024llama3} base LLMs due to computational limitations. The base model parameters remain frozen throughout, with domain-specific adaptation performed exclusively through lightweight LoRA modules. All inference experiments are conducted on V100 32G GPUs, with LLaMA-3-8B inference performed in bfloat16 precision format for computational efficiency.

\subsection{Baselines}
We compare our approach with different baselines spanning three fusion paradigms.

\noindent \textbf{Static fusion} is a naive baseline 
that assigns equal weightage to each participating LoRA without considering the relevance of respective adapters to the query at hand. 
This approach lacks query-adaptability and layer-level granularity.

\noindent \textbf{Supervised methods} learn optimal fusion weights from composite data. LoRAFlow~\cite{wang2024loraflow} trains a parametric router using composite examples per adapter combination to predict fusion weights. LoRAHub~\cite{huang2023lorahub} performs gradient-free optimization of fusion weights. Although effective, these methods heavily rely on training data for different adapter combinations, thus lacking scalability and generalizability.

\noindent \textbf{Training-free methods} like ~\cite{belofsky2023_cosine, chronopoulou2023adaptersoup} avoid supervised optimization of fusion weights by computing domain centroids from representative examples (subset of data used for training LoRA adapters), and assigning fusion weights based on respective cosine similarities. Although unsupervised, these approaches still require access to domain-representative data and do not capture the per-layer distributional shift introduced by adapters.

\noindent \textbf{Data and Training free methods} To the best of our knowledge, our method is the first under the data and training free paradigm. We differ from existing training-free methods by (i) eliminating dependence on representative examples entirely, and (ii) utilizing dynamic fusion weights per-layer.

\subsection{Datasets}

Our objective is to investigate the effectiveness of different LoRA fusion methods in handling challenging composite queries, where multi-domain expertise is intricately amalgamated in a query, rather than appearing as sequential tasks~\cite{xu2024meteora}. To this end, we draw inspiration from the evaluation tasks used in LoRAFlow~\cite{wang2024loraflow}. LoRAFlow evaluates fusion performance on six composite tasks combining three language adapters (Chinese, Russian, Spanish) with two domain adapters (Math, Code). 
To introduce more diversity in evaluation tasks, we extend their evaluation framework by introducing a Medical domain adapter, thereby enabling evaluation on nine multilingual composite tasks that span mathematical reasoning, code generation, and medical question answering, and require combining linguistic and domain expertise. In this section, we provide details on the datasets used for (i) training each LoRA expert, (ii) fusion weight optimization in supervised baselines, and (iii) our evaluation benchmarks. 

\subsubsection{LoRA expert training.}
To evaluate LoRA fusion performance, the six LoRA expert modules are trained as follows. The (i) Chinese (zh), (ii) Russian (ru), and (iii) Spanish (es) language experts are trained using the respective 52K conversational examples from~\cite{lai2023okapi}. The (iv) Math adapter is trained on 395K english mathematical reasoning problems from the MetaMathQA dataset~\cite{yu2023metamath}, the (v) Code adapter employs 186K english code generation problems from the MagiCoder dataset~\cite{wei2023magicoder}, and the (vi) Medical adapter is trained using 182K multiple-choice medical question-answer pairs from the MedMCQA dataset~\cite{ankit2022medical}.

All adapters except code are trained using a LoRA rank r=64 with scaling factor $\alpha$=16. Following~\cite{wang2024loraflow}, the code LoRA is trained using a rank r=256. Each LoRA adapter is trained for 3 epochs with a cosine warmup scheduling, where the peak learning rate is 1e-4, and the warmup ratio is 0.04.

\subsubsection{Data for supervised baselines.}  
Supervised LoRA fusion methods such as LoRAFlow~\cite{wang2024loraflow} and LoRAHub~\cite{huang2023lorahub} require training data for each task to learn optimal fusion weights. Towards this, for math and code tasks, we utilize the translated datasets from~\cite{wang2024loraflow}, which comprise 200 training examples for each of the six tasks. For medical tasks, we construct the training datasets by translating 280 medical QA examples from~\cite{ankit2022medical} into Chinese, Russian, and Spanish using GPT-4o with subsequent human verification. We follow the default training configurations of LoRAFlow~\cite{wang2024loraflow} and LoRAHub~\cite{huang2023lorahub} and train them on two A100 80G GPUs to benchmark their performance for our tasks. 

\subsubsection{Evaluation benchmarks.}  
To evaluate the 
fusion performance of different baselines in math tasks, we use 250 test samples from the MGSM dataset~\cite{shi2022math} which provides grade-school multilingual mathematical reasoning problems. 
For code tasks, we utilize 164 translated codes from the HumanEval dataset. For medical evaluation, we translate 150 test samples from the MedMCQA dataset~\cite{ankit2022medical} using GPT-4o with human verification. 

\subsection{Evaluation Metrics}
We employ different evaluation metrics for each domain. For mathematical reasoning tasks, we extract numerical answers from model outputs using regex-based postprocessing and compute accuracy against ground truth answers. Code generation tasks are assessed using the pass@1 metric, which measures the percentage of problems in which the generated code passes all test cases on the first attempt. Medical QA tasks use exact match scoring, where we evaluate whether the model's selected option matches the ground-truth answer.

\subsection{Results and Discussion}
\label{subsec:results}
\subsubsection{Quantitative analysis.} 
Table~\ref{tab:main_results} presents a comprehensive quantitative comparison of different LoRA fusion methods utilizing LLaMA-2-7B and LLaMA-3-8B as base LLMs, across nine composite tasks spanning mathematics, coding, and medical domains. 
Our proposed method \model{} 
substantially outperforms static and training-free baselines while significantly closing the gap with supervised baselines.

\begin{figure*}[!t]
\centering
\includegraphics[width=\textwidth]{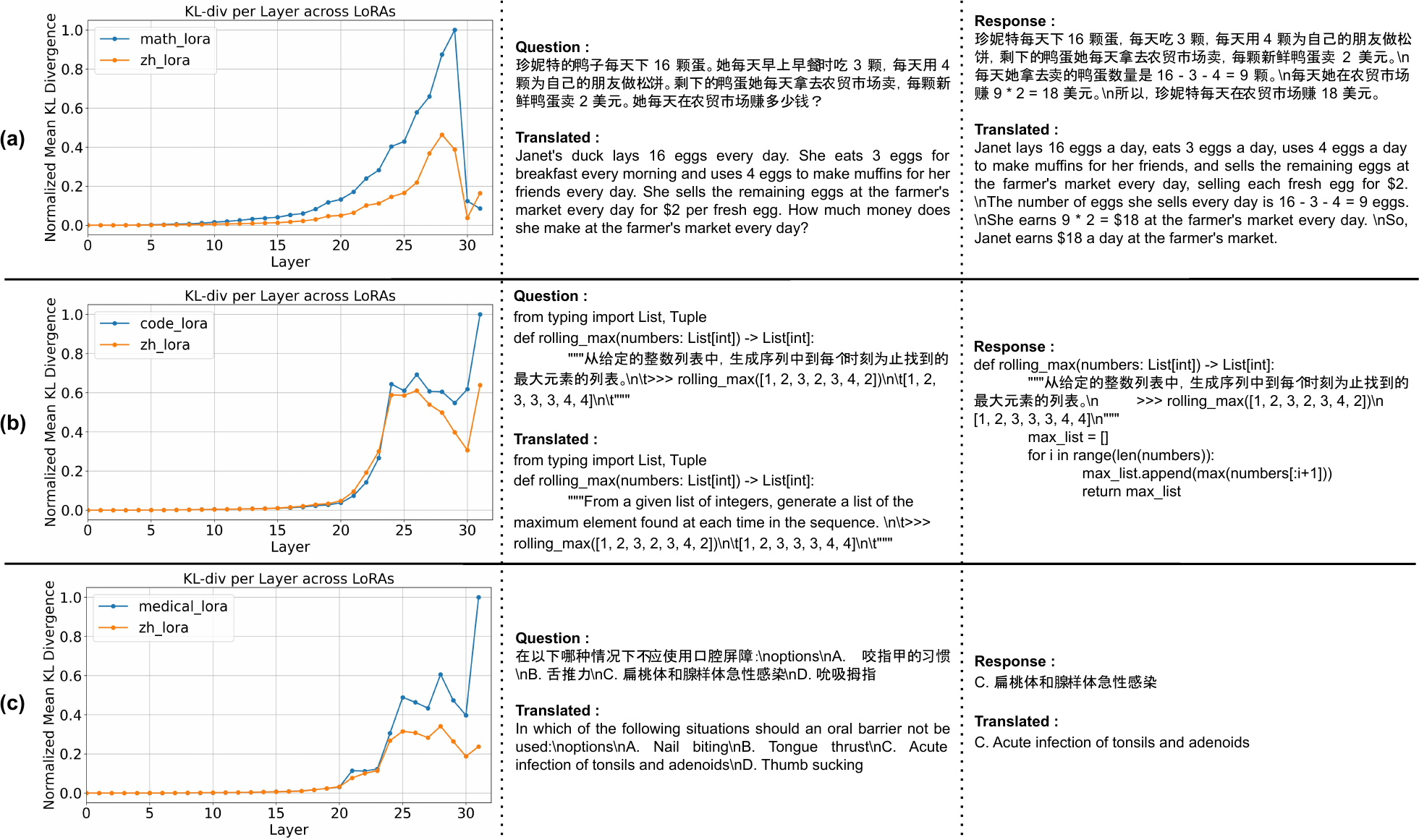}
\caption{\textbf{Layer-wise KL divergence analysis.} In the first column, we visualize the layer-level variation in mean KL divergence values (averaged across all test queries and then normalized) with LLaMA-2-7B base LLM for 3 composite tasks: (a) Chinese(zh)-math, (b) Chinese(zh)-code, and (c) Chinese(zh)-medical. The second and third columns show an example question-response pair (translations provided for understanding) for each of the three tasks.}
\label{fig:graphs}
\end{figure*}

Compared to the centroid similarity training-free baseline~\cite{belofsky2023_cosine, chronopoulou2023adaptersoup}, \model{} demonstrates superior overall average performance, achieving an improvement of $\sim$7\% with LLaMA-2 and $\sim$10\% with LLaMA-3 base LLM. This improvement is particularly pronounced in the mathematics domain, where \model{} outperforms centroid approach by $\sim$16\% with LLaMA-2 and $\sim$18\% with LLaMA-3. Similarly, the medical domain achieves an average improvement of $\sim$6\% with LLaMA-2 and $\sim$10\% with LLaMA-3. However, in the coding domain, both \model{} and centroid approach achieve comparable performance. This domain-specific performance variation can be explained as follows: Math and medical queries are language-heavy as illustrated in the second column of Figure~\ref{fig:graphs}. Centroid method overweighs language LoRA via lexical similarity, while \model{} overweighs task LoRA via distributional divergence. Thus, the centroid method has a lower performance in math and medical. Code queries on the other hand have both language(zh/ru/es) dominance and programming keywords (refer to second column of Figure~\ref{fig:graphs}). The lexical similarity measure in the centroid method causes higher weights for task LoRA due to keywords and syntactic matches. Thus, both methods produce similar fusion weights resulting in comparable performance.

The static fusion baseline, which naively employs equal weighing of adapters, achieves an overall average of 20.4\% with LLaMA-2 and 38.5\% with LLaMA-3. In contrast, our method delivers an improvement of $\sim$5\% with LLaMA-2 and $\sim$6\% with LLaMA-3. The consistent superior performance of \model{} over static fusion across all domains highlights the importance of dynamic, query-aware fusion weights. 

Supervised fusion approaches 
LoRAFlow~\cite{wang2024loraflow} and LoRAHub~\cite{huang2023lorahub} outperform our training-free method by 5.1\% and 0.6\% respectively with LLaMA-2 base LLM. Notably, with LLaMA-3, the performance gap with supervised LoRAFlow narrows significantly to just 1.9\%, suggesting that our approach scales effectively with more capable base models. Further, it is important to note that these supervised methods require a training phase to optimize the fusion weights. In contrast, our proposed method operates in a training‐free paradigm and even omits the requirement of representative samples (as in the centroid-based approach). \model{}'s ability to approach supervised performance while maintaining the flexibility and efficiency of data-and-training-free operation represents a significant practical advantage, especially when quality fusion data is expensive to obtain and repeated training for new adapter combinations is cumbersome. 

\subsubsection{Qualitative analysis.} 
To gain deeper insights into the fusion behavior of our method, we conduct a layer-level divergence analysis that reveals how respective domain and language adapters contribute across different network depths. Figure~\ref{fig:graphs} presents the KL divergence values of respective domain LoRAs and the Chinese language LoRA, averaged and normalized across all test queries for three composite tasks. 

We observe a consistent pattern in the initial transformer layers of the LLM, where KL divergence values approach zero for both domain and language adapters. This phenomenon aligns with established findings that lower transformer layers typically handle universal linguistic features~\cite{liu2024llm_semantics} that are well-captured during large-scale pre-training of the base LLM, requiring negligible task-specific adaptation.

In the Chinese(zh)-math task (Figure~\ref{fig:graphs}a), the math LoRA exhibits consistently higher KL divergence values throughout the middle layers (layers 10-30), reflecting its dominant role in foundational reasoning and arithmetic computations. However, there is a notable increase in the contribution from the Chinese LoRA in the final layer. This can be attributed to the generation phase: although the reasoning chain is mathematical, the final solution must be articulated in fluent Chinese with appropriate explanations and formatting. Thus, the language adapter becomes crucial for producing coherent, linguistically accurate responses that maintain mathematical precision while adhering to Chinese linguistic conventions.

\begin{table*}[t!]

\centering
\resizebox{1.0\textwidth}{!}{
\scriptsize
 \renewcommand{\arraystretch}{0.8}
\begin{tabular}{l|ccc|c|ccc|c|ccc|c|c}
\toprule
 \multirow{2}{*}{\textbf{Token Granularity}} & \multicolumn{4}{c|}{\textbf{Math}} & \multicolumn{4}{c|}{\textbf{Code}} & \multicolumn{4}{c|}{\textbf{Medical}} & \multirow{2}{*}{\makecell{\textbf{Avg across} \\ \textbf{3 domains}}} \\
 
\cmidrule(lr){2-5} \cmidrule(lr){6-9} \cmidrule(lr){10-13}

 & \textbf{zh} & \textbf{ru} & \textbf{es} & \textbf{Avg} & \textbf{zh} & \textbf{ru} & \textbf{es} & \textbf{Avg} & \textbf{zh} & \textbf{ru} & \textbf{es} & \textbf{Avg} & \\
 
\cmidrule(lr){1-1} \cmidrule(lr){2-5} \cmidrule(lr){6-9} \cmidrule(lr){10-13} \cmidrule(lr){14-14}

 Full query & 18.8 & 18.4 & 26.4 & 21.2 & \textbf{20.9} & \textbf{18.0} & \textbf{17.7} & \textbf{18.9} & 27.3 & 36.0 & 27.9 & 30.4 & 23.5 \\
 Last token (Ours) & \textbf{21.6} & \textbf{21.6} & \textbf{36.4} & \textbf{26.5} &\textbf{ 20.9} & 16.5 & 15.6 & 17.7 & \textbf{30.0} & \textbf{39.0} & \textbf{31.0} & \textbf{33.3} & \textbf{25.8} \\
 
\bottomrule
\end{tabular}}
\caption{Ablation Study to identify the optimal token-level granularity for best performance.}
\label{tab:ablation_token}
\end{table*}

In the Chinese(zh)-code task (Figure~\ref{fig:graphs}b), we observe a slight dominance of Chinese LoRA in the middle layers(20-23). This phase corresponds to the interpretation stage, where the model must fully comprehend the algorithmic requirements, constraints, and expected functionality described in Chinese (code comment). Following this interpretation phase, the code LoRA assumes dominant influence across all subsequent layers, reflecting the transition from language understanding to code synthesis. The generation process involves universal programming language constructs (keywords, operators, control structures) that are language-agnostic. Once the initial intent is decoded from the Chinese description, the subsequent generation process relies heavily on the code adapter's specialized knowledge of programming patterns, algorithmic structures, and syntax rules.

The Chinese(zh)-medical task (Figure~\ref{fig:graphs}c) demonstrates a medical-dominant pattern, where the domain adapter maintains higher KL divergence values, particularly evident across the final layers. This pattern reflects the requirement for precise terminology understanding in the middle layers and comprehensive medical domain knowledge to enable choosing the correct option in the final layers. The persistent high contribution of medical LoRA throughout the network depth indicates that medical reasoning requires continuous access to specialized knowledge, including disease pathophysiology, diagnostic criteria and treatment protocols.

The interpretability provided by these layer-level visualizations serves as both a theoretical validation of our method's effectiveness and a diagnostic tool for understanding fusion dynamics. 
This analysis suggests that an optimal fusion strategy must 
capture the layer-level dynamics of how different expertise are required at different processing stages of a query. Appendix A also provides a similar layer-level analysis for the remaining composite tasks.

\subsubsection{Ablation Study: Optimal query tokens granularity for relevance estimation.}
We investigate the optimal token granularity for KL divergence computation by comparing two approaches: averaging divergence across all query tokens versus using only the last token's divergence. Table~\ref{tab:ablation_token} shows that the latter approach outperforms all-token averaging by $\sim$2\%. This performance gap can be attributed to the autoregressive nature of transformer models, where the final token's hidden state encapsulates the full sequential context through self-attention mechanisms. Additionally, relying on the last-token alone reduces computational overhead by eliminating position-wise calculations, making it both effective and efficient for adapter relevance estimation.

Appendix A discusses another ablation study justifying the choice of our divergence measure technique.

\subsubsection{Latency Analysis.} 
To evaluate the computational efficiency of our approach, we measure the average latency for 250 queries of the Chinese(zh)-math task using LLaMA-2-7B base LLM. The queries average 154 tokens in length. We perform all evaluations on V100 32G GPUs.

The inference process comprises two components: (a) fusion weight computation, which adds $\sim$192ms per query per adapter. This overhead stems from the forward passes required to extract layer-level hidden states and their probability distributions to compute KL divergences. Importantly, this computation can be parallelized across adapters, enabling substantial speedup. (b) generation time, which remains comparable to supervised LoRAFlow method.

While \model{} introduces a negligible overhead for fusion-weight computation, it completely eliminates the training phase required by supervised methods. Thus, there is no need for composite data collection and fusion weights optimization for all possible adapter combinations. Our training-free paradigm computes fusion weights on-the-fly, making it readily applicable to new adapter collections and substantially more scalable as the number of adapters grow.

\section{Conclusion}
In this work, we propose \model{}, a novel training-free approach for query-adaptive LoRA fusion that dynamically integrates multiple domain-specific adapters. Our method leverages distributional divergence between adapter and base model representations at each layer, to quantify the semantic relevance of each adapter to the query, thereby enabling principled and interpretable fusion weight computation. Extensive experimental evaluation across nine composite tasks demonstrates that \model{} achieves substantial improvements, outperforming static and training-free methods by large margins, while closing the gap with supervised fusion approaches that require additional training overhead. Overall, our approach offers a scalable and effective solution for training-free adapter fusion, eliminating the need for additional composite data, and setting a strong foundation for future research in unsupervised adapter fusion techniques.

\section{Limitations and Future Work}
Our evaluation is restricted to the LLaMA-2-7B and LLaMA-3-8B models due to computational constraints. While we demonstrate improvements across nine diverse composite tasks, future work could further validate our approach with varied-scale LLMs (13B, 70B variants). 

Despite achieving substantial improvements over training-free baselines, our method still exhibits a performance gap compared to supervised fusion approaches, particularly in domains requiring complex reasoning. Future research could explore more sophisticated relevance measures beyond KL divergence, while preserving the training-free paradigm. Moreover, investigating fusion strategies that can dynamically select between different relevance measures based on query characteristics represents a promising avenue to close the remaining performance gap with supervised methods.

\bibliography{aaai2026}


\clearpage
\appendix

\begin{center}
  \Large\bfseries Appendix
\end{center}

\section{Results and Discussion}

\subsection{Qualitative analysis of six composite tasks}

\begin{table*}[h!]
\centering
\resizebox{0.95\textwidth}{!}{
\scriptsize
\renewcommand{\arraystretch}{0.9}
\begin{tabular}{l|ccc|c|ccc|c|ccc|c|c}
\toprule
 \multirow{2}{*}{\textbf{Divergence measure}} & \multicolumn{4}{c|}{\textbf{Math}} & \multicolumn{4}{c|}{\textbf{Code}} & \multicolumn{4}{c|}{\textbf{Medical}} & \multirow{2}{*}{\makecell{\textbf{Avg across} \\ \textbf{3 domains}}} \\
 
\cmidrule(lr){2-5} \cmidrule(lr){6-9} \cmidrule(lr){10-13}

 & \textbf{zh} & \textbf{ru} & \textbf{es} & \textbf{Avg} & \textbf{zh} & \textbf{ru} & \textbf{es} & \textbf{Avg} & \textbf{zh} & \textbf{ru} & \textbf{es} & \textbf{Avg} & \\
 
\cmidrule(lr){1-1} \cmidrule(lr){2-5} \cmidrule(lr){6-9} \cmidrule(lr){10-13} \cmidrule(lr){14-14}

 Cosine distance & 20.4 & \textbf{22.4} & 36.0 & 26.3 & \textbf{21.7} & 15.9 & 17.1 & 18.2 & 29.0 & 38.0 & \textbf{32.0} & 33.0 & \textbf{25.8} \\
 Euclidean distance & 18.4 & 19.2 & 27.2 & 21.6 & \textbf{21.7} & 15.5 & \textbf{17.7} & \textbf{18.3} & \textbf{31.0} & 37.0 & 28.0 & 32.0 & 24.0 \\
 KL divergence (Ours) & \textbf{21.6} & 21.6 & \textbf{36.4} & \textbf{26.5} & 20.9 & \textbf{16.5} & 15.6 & 17.7 & 30.0 & \textbf{39.0} & 31.0 & \textbf{33.3} & \textbf{25.8} \\
 
\bottomrule
\end{tabular}}
\caption{Ablation Study to quantify the impact of the choice of divergence measure for adapter relevance estimation.}
\label{tab:ablation_distance}
\end{table*}

\begin{figure*}[!t]
\centering
\includegraphics[width=0.98\textwidth]{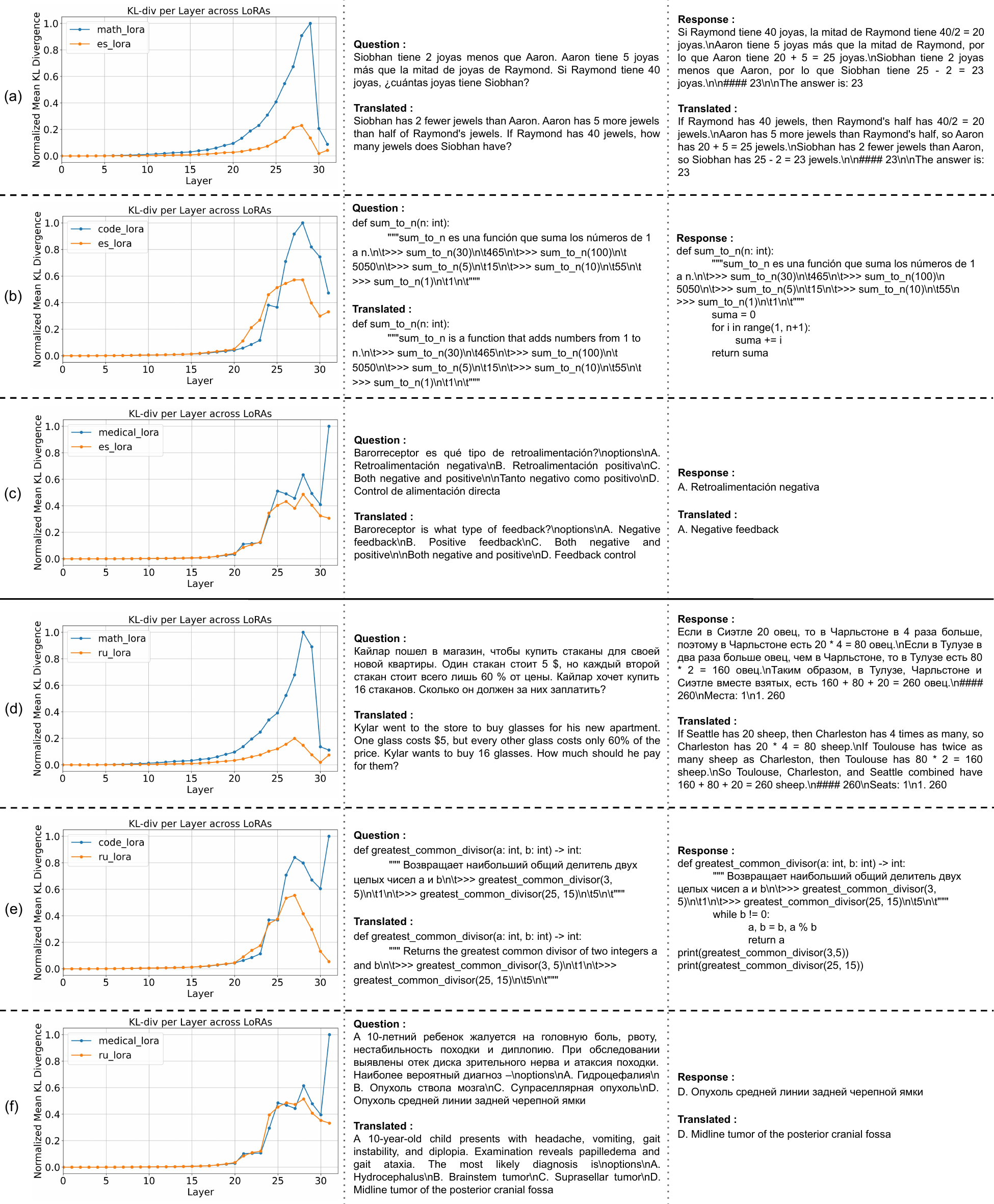}
\caption{\textbf{Layer-wise KL divergence analysis.} In the first column, we visualize the layer-level variation in mean KL divergence values (averaged across all test queries and then normalized) with LLaMA-2-7B base LLM for 6 composite tasks (a) Spanish(es)-math, (b) Spanish(es)-code, (c) Spanish(es)-medical, (d) Russian(ru)-math, (e) Russian(ru)-code, and (f) Russian(ru)-medical. The second and third columns show an example question-response pair (translations provided for understanding) for each of the six tasks.}
\label{fig:graphs_rem}
\end{figure*}

To provide comprehensive insights into our fusion mechanism, we extend the layer-wise divergence analysis to the remaining six composite tasks involving Spanish and Russian languages. Figure~\ref{fig:graphs_rem} presents the KL divergence patterns for these tasks, revealing both consistent trends and language-specific variations in adapter contributions.

Consistent with our analysis of Chinese tasks, all composite tasks exhibit zero KL divergence in initial layers, confirming that lower transformer layers capture universal linguistic features requiring minimal task-specific adaptation.

For mathematical reasoning tasks (Figure~\ref{fig:graphs_rem} a,d), both Spanish and Russian variants show sustained math LoRA dominance throughout the middle and final layers, contrasting with the Chinese task where the language adapter contribution increases in the final layer. This difference reflects the linguistic proximity of Spanish and Russian to English (the base model's primary training language), requiring less language-specific adaptation for generating mathematical explanations compared to Chinese.

The coding tasks (Figure~\ref{fig:graphs_rem} b,e) demonstrate similar interpretation-to-synthesis transitions as observed with Chinese, where language adapters dominate during the problem comprehension phase (layers 20-23) before the code adapter assumes control for universal programming construct generation. This consistency validates that our method captures the fundamental cognitive processing stages.

Medical tasks (Figure~\ref{fig:graphs_rem} c,f) maintain a domain-dominant pattern throughout all six tasks, with medical LoRA exhibiting high divergence values through the middle and final layers. This consistency reinforces that medical reasoning demands continuous access to specialized domain knowledge.

The interpretability provided by this layer-level analysis across all nine composite tasks serves as both a theoretical validation of our method's effectiveness and a diagnostic tool for understanding fusion dynamics. \\

\subsection{Ablation Study: Choice of divergence measure for adapter relevance estimation} 
We evaluate three distance metrics to quantify adapter relevance: KL divergence on vocabulary distributions (our approach), cosine distance on hidden states (where we define cosine\_distance = 1-cosine\_similarity), and euclidean distance on hidden states. As shown in table~\ref{tab:ablation_distance}, our approach utilizing KL divergence yields the highest average performance across tasks, similar to cosine distance (25.8\%) and surpassing euclidean distance (24\%). 

While geometric distances like cosine similarity and euclidean distance capture representational similarity by operating in hidden state space, KL divergence operates in probability space directly reflecting the model's predictive behavior and confidence. This probabilistic formulation provides a more principled adapter relevance estimation by measuring distributional differences rather than geometric proximity.

\end{document}